\documentclass[10pt,twocolumn,letterpaper]{article}

\usepackage{iccv}
\usepackage{times}
\usepackage{epsfig}
\usepackage{graphicx}
\usepackage{amsmath}
\usepackage{amssymb}

\usepackage[linesnumbered,ruled,vlined]{algorithm2e}

\usepackage{pifont}
\newcommand{\cmark}{\ding{51}}%
\usepackage{makecell}
\usepackage{multirow}
\usepackage{booktabs}
\usepackage{enumerate}
\newcommand{\PreserveBackslash}[1]{\let\temp=\\#1\let\\=\temp}
\newcolumntype{C}[1]{>{\PreserveBackslash\centering}p{#1}}
\newcolumntype{R}[1]{>{\PreserveBackslash\raggedleft}p{#1}}
\newcolumntype{L}[1]{>{\PreserveBackslash\raggedright}p{#1}}
\usepackage[pagebackref=true,breaklinks=true,letterpaper=true,colorlinks,bookmarks=false]{hyperref}

\newcommand{\tabincell}[2]{\begin{tabular}{@{}#1@{}}#2\end{tabular}}

\newcommand{\HF}[1]{\textcolor[rgb]{0.00,0.00,0.00}{#1}}

\newcommand{\supp}[1]{\textcolor[rgb]{0.00,0.00,0.00}{#1}}

\iccvfinalcopy 

\def\iccvPaperID{****} 

\renewcommand\footnotemark{}

\begin{document}
	
	\title{Transparent Object Tracking Benchmark}
	
	\author{Heng Fan$^{1}$ \;\; Halady Akhilesha Miththanthaya$^{2*}$ \;\; Harshit$^{2*}$ \;\; Siranjiv Ramana Rajan$^{2*}$ 
		\\ Xiaoqiong Liu$^{2}$ \;\; Zhilin Zou$^{2}$ \;\; Yuewei Lin$^{3}$ \;\; Haibin Ling$^{2\dag}$ \\
		$^{1}$Department of Computer Science and Engineering, University of North Texas, Denton, USA\\
		$^{2}$Department of Computer Science, Stony Brook University, Stony Brook, USA\\
		$^{3}$Computational Science Initiative, Brookhaven National Laboratory, Upton, USA\\
		{\tt\small heng.fan@unt.edu \;\;\; hling@cs.stonybrook.edu}
		\thanks{$^{*}$The three authors make equal contributions.}
		\thanks{$^{\dag}$Corresponding author.}
		}
	
	\maketitle
	
	\begin{abstract}
		Visual tracking has achieved considerable progress in recent years. However, current research in the field mainly focuses on tracking of opaque objects, while little attention is paid to transparent object tracking. In this paper, we make the first attempt in exploring this problem by proposing a Transparent Object Tracking Benchmark (TOTB). Specifically, TOTB consists of 225 videos (86K frames) from 15 diverse transparent object categories. Each sequence is manually labeled with axis-aligned bounding boxes. To the best of our knowledge, TOTB is the first benchmark dedicated to transparent object tracking. In order to understand how existing trackers perform and to provide comparison for future research on TOTB, we extensively evaluate 25 state-of-the-art tracking algorithms. The evaluation results exhibit that more efforts are needed to improve transparent object tracking. Besides, we observe some nontrivial findings from the evaluation that are discrepant with some common beliefs in opaque object tracking. For example, we find that deeper features are not always good for improvements. Moreover, to encourage future research, we introduce a novel tracker, named TransATOM, which leverages transparency features for tracking and surpasses all 25 evaluated approaches by a large margin. By releasing TOTB, we expect to facilitate future research and application of transparent object tracking in both the academia and industry. The TOTB and evaluation results as well as TransATOM are available at \url{https://hengfan2010.github.io/projects/TOTB/}.
	\end{abstract}
	
	\section{Introduction}
	
	\begin{figure}[!t]
		\centering
		\includegraphics[width=\linewidth]{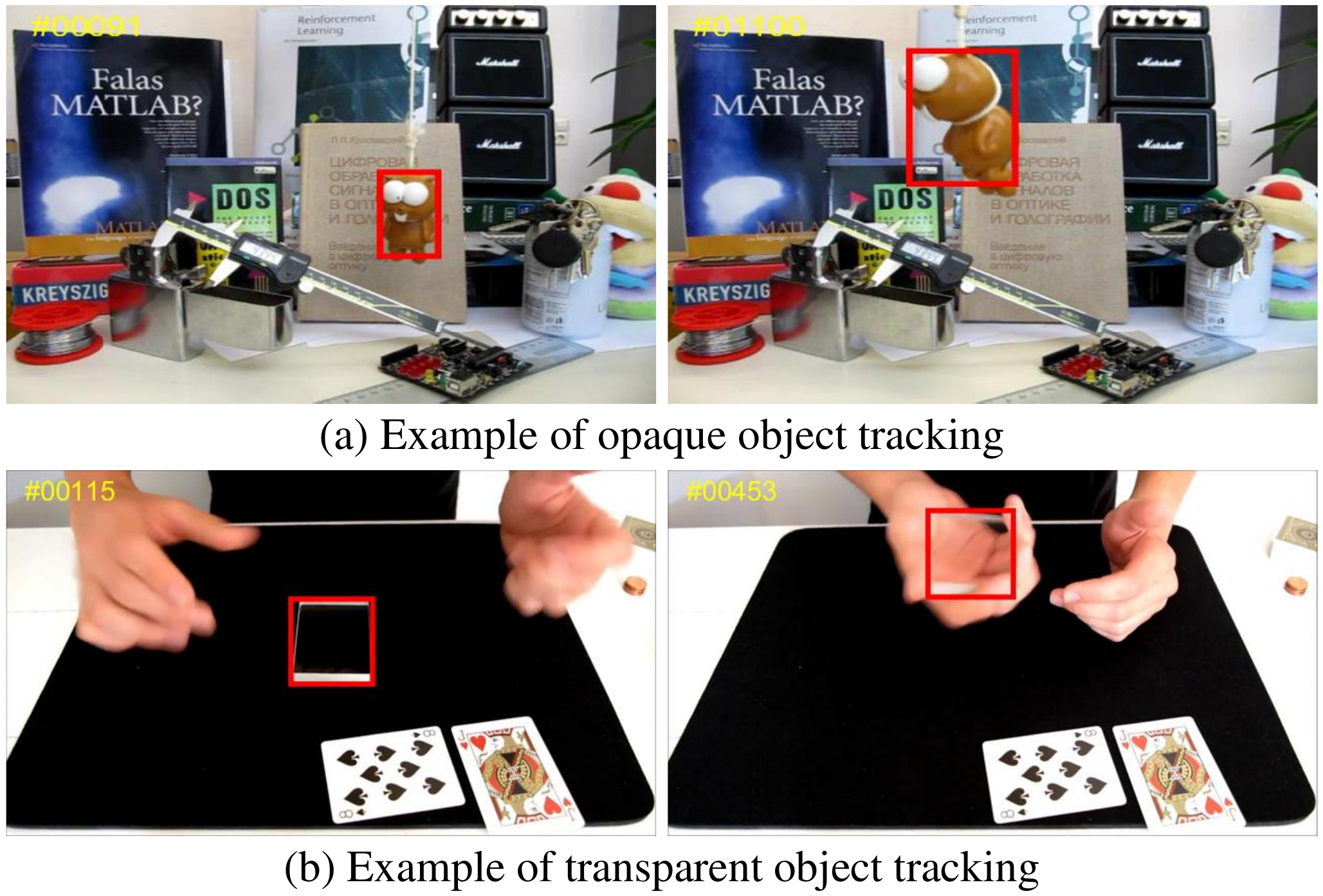}
		\caption{Opaque object tracking (a) and transparent object tracking (b). Compared with opaque object tracking in which target object appearance is more distinguishable from background and consistent over time, tracking of transparent target is more challenging as transparent object appearance is heavily dependent on background. \emph{All figures in this paper are best viewed in color and by zooming in}.
		}
		\label{fig:comparison}
	\end{figure}

	Object tracking is one of the most fundamental problems in computer vision and serves as an important component in numerous applications~\cite{li2013survey,smeulders2014visual,yilmaz2006object,li2018deep} including robotics, human-machine interaction, video analysis and understanding, \etc. In recent decades, the tracking community has witnessed remarkable progress. Numerous tracking algorithms 
	have been proposed and significantly pushed the state-of-the-arts. Nevertheless, existing research in the field mainly focuses on opaque object tracking, while \emph{very little} attention is paid to tracking of transparent objects. 
	
	Transparent objects (\eg, \emph{bottle}, \emph{cup}, \emph{bulb}, \emph{jar} and many others made by glass and plastics) are common to see in the real world. Many of them are closely related to human daily life, and tracking of them are crucial for robotic vision and human-machine interaction. For example, a robot may need to know the trajectory of a transparent object in human hand for better action understanding.
	
	Compared with tracking of opaque objects, transparent object tracking is more challenging. Because of the particular \emph{transparency} feature, the appearances of transparent objects are relatively \emph{weak} and largely mixed with the surrounding background image (see Figure~\ref{fig:comparison} for an example). As a result, it becomes more difficult to directly leverage appearance information to distinguish the target object from background. In addition, when a target object moves, even slowly, its appearance may change drastically due to background variation, making transparent object tracking harder.
	
	Besides the above technical difficulty, another more important reason that transparent object tracking is untouched is because of lack of a benchmark. Benchmark is crucial for the advancement of tracking. It allows researchers to objectively evaluate and compare their methods as well as design new algorithms for improvement. Currently, there exist various benchmarks (\eg,~\cite{wu2015object,mueller2016benchmark,galoogahi2017need,muller2018trackingnet,valmadre2018long,kristan2018visual,fan2019lasot,huang2018got,lukezic2019cdtb}) for opaque object tracking. However, there is \emph{no} benchmark for transparent object tracking. Although some of benchmarks (\eg,~\cite{fan2019lasot,liang2015encoding}) consist of sequences of transparent objects, they are limited in both number of videos (\eg, less than 10) and object classes (\eg, at most two categories). To facilitate research on transparent object tracking, a dedicated dataset is desired to serve as the testbed for fair evaluation and comparison.

	\subsection{Contribution}
	
	In this work, we make the \emph{first} attempt in exploring transparent object tracking by introducing a Transparent Object Tracking benchmark (TOTB), \HF{which is our major contribution}. TOTB comprises of a diverse selection of 15 common transparent object classes with each containing 15 sequences. In total, TOTB consists of 225 sequences with 87K frames. Each sequence is manually annotated with axis-align bounding boxes and labeled with different attributes. To our best knowledge, TOTB is the \emph{first} benchmark dedicated to the task of transparent object tracking. Figure~\ref{fig1} demonstrates several example sequences in TOTB.
	
	Besides, in order to understand how existing tracking algorithms perform and to provide comparisons for future research on TOTB, we extensively evaluate 25 state-of-the-art trackers. We conduct in-depth analysis on the evaluation results and observe several surprising findings that are discrepant with some popular beliefs in the opaque object tracking. For example, it is widely believed that deeper features are crucial to improve tracking performance, as shown in the existing opaque tracking benchmarks (\eg,~\cite{wu2015object,fan2019lasot,muller2018trackingnet,huang2018got}). Contrary to this, it turns out that deeper features do not always bring performance gains for transparent object tracking. Instead, it may heavily decrease accuracy. These observations provide better understanding of transparent object tracking and guidance for future improvements.
	
	Furthermore, to facilitate the development of tracking algorithms on TOTB, we introduce a {\it simple yet effective} tracker by exploiting transparency features for tracking. In particular, considering that transparency is a common attribute of transparent objects, its feature should be generic and transferable for all transparent instances, and also distinguishable from opaque objects. To this end, we train a deep network to learn such transparency feature and apply it for tracking by integrating it into ATOM~\cite{danelljan2019atom}. Our new tracker, dubbed TransATOM, is assessed on TOTB and significantly outperforms all evaluated algorithms by a large margin. \HF{Note that, although TransATOM is simple, it demonstrates the effectiveness of transparency feature in boosting performance. We expect it to provide a reference for facilitating future study.}
	
	In summary, we make the following contributions:
	
	\noindent
	{\bf (1)} \emph{We propose TOTB, which is, to the best of knowledge, the first benchmark dedicated for transparent object tracking.}
	
	\noindent
	{\bf (2)} \emph{To assess existing trackers and provide comparisons, we evaluate 25 tracking algorithms with in-depth analysis.}
	
	\noindent
	{\bf (3)} \emph{We introduce a novel transparent object tracker, named TransATOM, to encourage further research on TOTB.}

	By releasing TOTB, we hope to facilitate future research and application of transparent object tracking.
	
	The rest of this paper is organized as follows. We discuss related works of this paper in Section~\ref{relatedwork}. Section~\ref{TOTB} details the proposed TOTB. Section~\ref{transatom} introduces our proposed tracker TransATOM. Evaluation results are shown in Section~\ref{res} with in-depth analysis, followed by conclusion in Section~\ref{con}.

	\section{Related Work}
	\label{relatedwork}
	
	\subsection{Tracking Algorithm}
	
	As one of core members in the computer vision family, visual tracking has been studied for decades, with a huge past literature whose review is beyond this paper. In this section, we review two popular trends including correlation filter tracking and deep tracking in the field and refer readers to~\cite{li2013survey,smeulders2014visual,yilmaz2006object,li2018deep} for comprehensive tracking surveys.
	
	Roughly speaking, correlation filter-based tracking algorithms treat tracking as an online regression problem. Correlation filter trackers like~\cite{bolme2010visual,henriques2015high} demonstrate impressive running speeds of several hundreds frames per second and attract great attention in the tracking community with many inspired extensions for improvements. For example, an additional scale filter is utilized in~\cite{li2014scale,danelljan2014accurate} to deal with the target scale variations. The approaches in~\cite{danelljan2015learning,galoogahi2017learning,danelljan2017eco,li2018learning} leverage regularization techniques to improve robustness. The tracker in~\cite{fan2017parallel} integrates correlation filter tracker with an independent verifier to alleviate the drifting problem. The methods of~\cite{ma2015hierarchical,danelljan2016beyond,dai2019visual} apply deep features to replace hand-crafted ones in correlation filter tracking and achieve significant improvements.

	Motivated by the tremendous success of deep features in other vision tasks, deep learning-based trackers have been developed in recent years. Among them, a popular series follows the Siamese trackers~\cite{tao2016siamese,bertinetto2016fully}, which present a simple architecture yet promising performance. Notably, a fully convolutional Siamese network is introduced in~\cite{bertinetto2016fully} with a light structure for tracking, leading to very efficient running performance. Inspired by the balanced accuracy and speed of~\cite{bertinetto2016fully}, many other variants~\cite{he2018twofold,li2018high,li2019siamrpn++,li2019gradnet,zhu2018distractor,fan2019siamese,wang2019spm,zhang2019deeper,zhang2020ocean,fan2021cract} have been developed and generated boosted performances. Along another line, some deep trackers~\cite{danelljan2019atom,bhat2019learning,danelljan2020probabilistic} decompose tracking into two separate localization and scale estimation tasks, which are respectively solved by an online classifier and an offline intersection-over-union (IoU) network.

	\subsection{Tracking Benchmark}
	
	Benchmarks are crucial for the development of tracking. We roughly categorize existing benchmarks into two types: \emph{generic} benchmark and \emph{specific} benchmark.
	
	\noindent
	{\bf Generic Benchmark.} A generic tracking benchmark usually includes sequences for general scenes. OTB-2013~\cite{wu2015object} is the first generic dataset with 50 sequences and later extended in larger OTB-2015~\cite{wu2015object} by introducing extra videos. TC-128~\cite{liang2015encoding} collects 128 colorful sequences to investigate the impact of color information on tracking performance. VOT~\cite{kristan2016novel} introduces a series of tracking competitions with up to 60 sequences. NfS~\cite{galoogahi2017need} focuses on evaluating trackers on videos with high frame rate. NUS-PRO~\cite{li2016nus} offers 365 videos with the goal of performance evaluation on rigid objects. TracKlinic~\cite{fan2019tracklinic} provides 2,390 videos with a diagnosis goal of tracking algorithms under various challenges. Recently, to provide training data for developing deep trackers, many large-scale benchmarks have been proposed. OxUvA~\cite{valmadre2018long} provides 366 videos with the goal of long-term evaluation. TrackingNet~\cite{muller2018trackingnet} consists of more than 30K sequences for deep tracking. GOT-10k~\cite{huang2018got} offers 10K videos with rich motion trajectories for tracking. LaSOT~\cite{fan2019lasot} comprises 1,400 long-term videos with manual annotations. Later, LaSOT is extended in~\cite{fan2020lasot} by introducing 150 new video sequences and a new evaluation protocol for unseen objects with more analysis.
	
	\noindent
	{\bf Specific Benchmark.} Besides generic datasets, there exist other benchmarks for specific goals. UAV and UAV123~\cite{mueller2016benchmark} consists of 100 and 23 videos captured by unmanned aerial vehicle (UAV). CDTB~\cite{lukezic2019cdtb} and PTB~\cite{song2013tracking} aim at assessing tracking performance on RGB-D videos. VOT-TIR~\cite{kristan2017visual} is from VOT and focuses on object tacking in RGB-T sequences.
	
	Despite of the availability of the above benchmarks, they mainly focus on opaque object tracking. Tracking of transparent target objects, which widely appear in the real-world, has received \emph{very little} attention. The most important reason is the lack of a dataset for transparent object tracking, which motivates our proposal of TOTB.
	
	\subsection{Dealing with Transparent Object in Vision}
	
	Transparent objects are common to see in the real-world, and a significant amount of research has been devoted to deal with them. For example, the methods of~\cite{fritz2009additive,maeno2013light} investigate the problem of transparent object recognition. The approach of~\cite{klank2011transparent} explores the time of flight (ToF) camera to detect and reconstruct transparent objects. The approach of~\cite{liu2020keypose} proposes to estimate keypoints of transparent objects in RGB-D images. The work of~\cite{sajjan2020clear} studies the problem of 3D shape estimation for transparent objects in RGB-D images. The methods of~\cite{xu2015transcut,kalra2020deep,xie2020segmenting} handle the task of segmenting transparent objects from an image. Especially, the work of~\cite{xie2020segmenting} presents a large-scale benchmark for transparent object segmentation.

	Our work is related to~\cite{liu2020keypose,sajjan2020clear,xie2020segmenting} but different in: (1) TOTB focuses on 2D object tracking, while other works on 3D shape estimation~\cite{sajjan2020clear}, 3D labeling and keypoint estimation~\cite{liu2020keypose} and 2D object segmentation~\cite{xie2020segmenting}. (2) TOTB deals with transparent objects in videos, while \cite{liu2020keypose,sajjan2020clear,xie2020segmenting} in static images.

	\section{Transparent Object Tracking Benchmark}
	\label{TOTB}
	
	We aim to construct a dedicated transparent object tracking benchmark (TOTB). When developing TOTB, we cover a diverse selection of transparent object classes and provide manual annotations for each video, as detailed later.
	
	\begin{figure}[!t]
		\centering
		\includegraphics[width=0.18\linewidth, height=1.5cm]{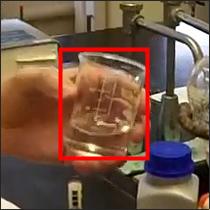}
		\includegraphics[width=0.18\linewidth, height=1.5cm]{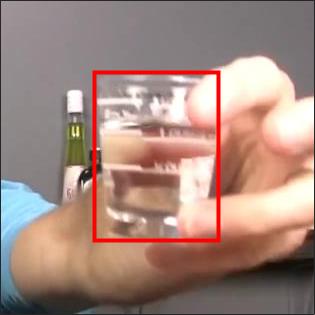}
		\includegraphics[width=0.18\linewidth, height=1.5cm]{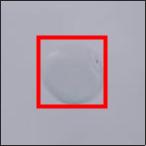}
		\includegraphics[width=0.18\linewidth, height=1.5cm]{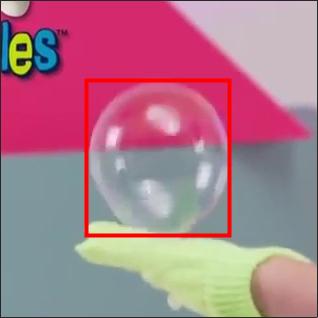}
		\includegraphics[width=0.18\linewidth, height=1.5cm]{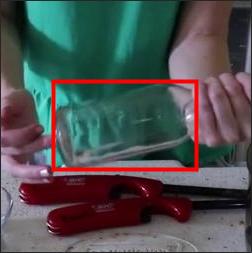} \\
		\includegraphics[width=0.18\linewidth, height=1.5cm]{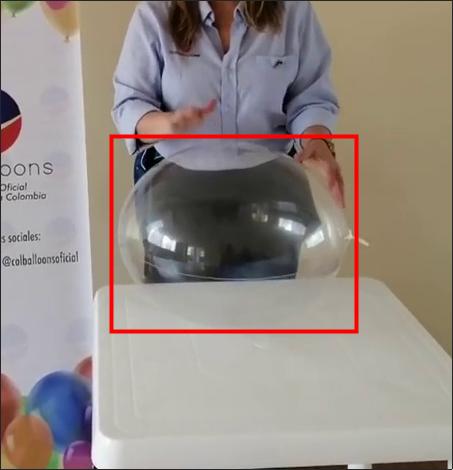}
		\includegraphics[width=0.18\linewidth, height=1.5cm]{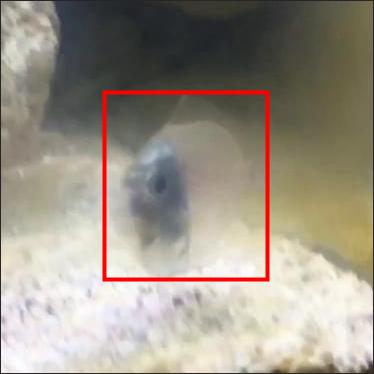}
		\includegraphics[width=0.18\linewidth, height=1.5cm]{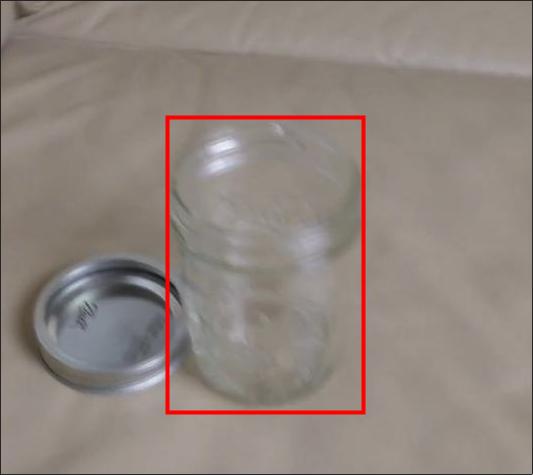}
		\includegraphics[width=0.18\linewidth, height=1.5cm]{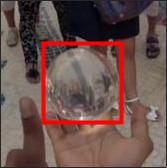}
		\includegraphics[width=0.18\linewidth, height=1.5cm]{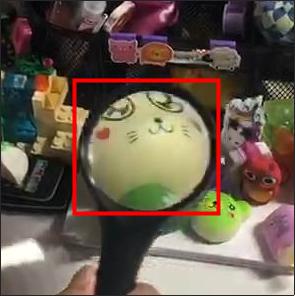} \\
		\includegraphics[width=0.18\linewidth, height=1.5cm]{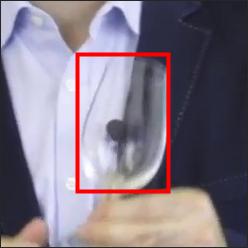}
		\includegraphics[width=0.18\linewidth, height=1.5cm]{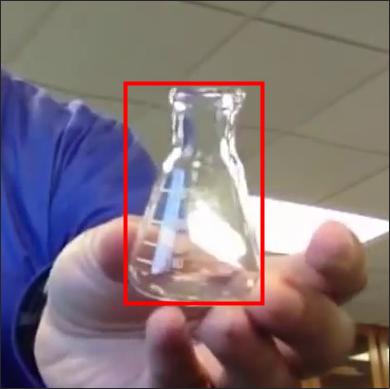}
		\includegraphics[width=0.18\linewidth, height=1.5cm]{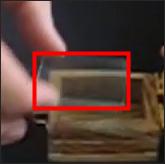}
		\includegraphics[width=0.18\linewidth, height=1.5cm]{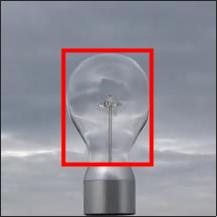}
		\includegraphics[width=0.18\linewidth, height=1.5cm]{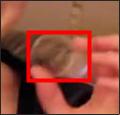}
		\caption{Samples from 15 transparent object categories. First row: Beaker, GlassCup, WubbleBubble, JuggleBubble and GlassBottle. Second row: BubbleBalloon, TransparentAnimal, GlassJar, GlassBall and MagnifyingGlass. Third row: WineGlass, Flask, GlassSlab, Bulb and ShotGlass. The tracking targets are shown in the red bounding boxes.}
		\label{fig_15_samples}
	\end{figure}
	
	\subsection{Video Collection}
	
	In TOTB, we select 15 transparent object categories consisting of \emph{Beaker}, \emph{GlassCup}, \emph{WubbleBubble}, \emph{JuggleBubble}, \emph{GlassBottle}, \emph{BubbleBalloon}, \emph{TransparentAnimal}, \emph{GlassJar}, \emph{GlassBall}, \emph{MagnifyingGlass}, \emph{WineGlass}, \emph{Flask}, \emph{GlassSlab}, \emph{Bulb} and \emph{ShotGlass}. Note that, the transparent \emph{window} and \emph{door} widely appear in the real-world, nevertheless, the objects of these two categories are usually static, and therefore not suitable for tracking task. Figure~\ref{fig_15_samples} demonstrates the samples from these 15 categories.
	
	\renewcommand\arraystretch{1}
	\begin{table}[!t]\small
		\centering
		\caption{Summary of statistics of the proposed TOTB. OV: out-of-view; FOC: full occlusion.}
		\begin{tabular}{rl|rl}
			\Xhline{3\arrayrulewidth}
			Number of videos & 225   & Avg. duration & 12.7s \\
			Total frames & 86$\mathbf{K}$ ~~   & Frame rate & 30 \emph{fps} \\
			Max frames & 500   & Absent labels & OV, FOC \\
			Min frames & 126   & ~Object categories & 15 \\
			Avg. frames & 381   & Number of att. & 12 \\
			\Xhline{3\arrayrulewidth}
		\end{tabular}%
		\label{tab:totb}%
	\end{table}%
	
	\begin{figure}[!t]
		\centering
		\includegraphics[width=\linewidth]{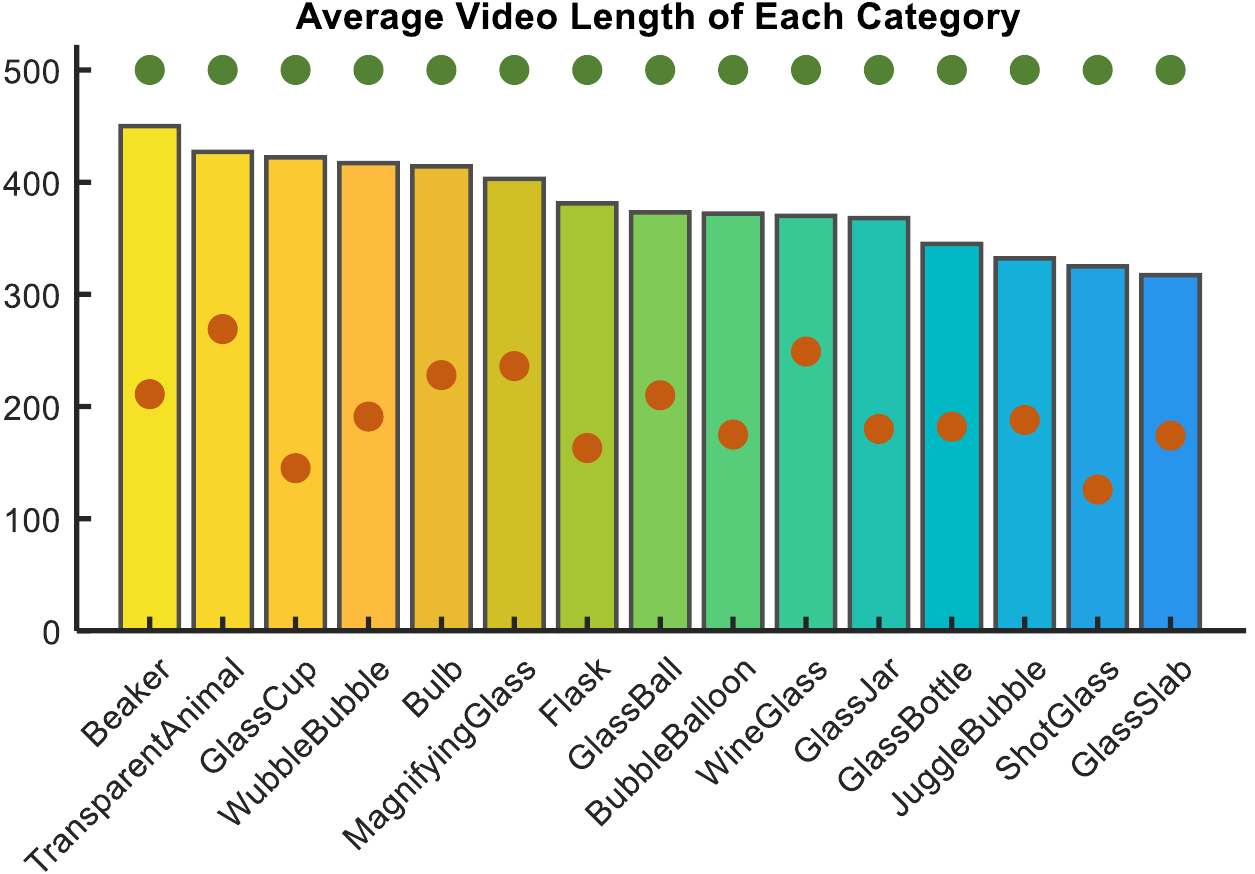}
		\caption{Average video length for each object class in TOTB. The green and brown dots represent the maximum and minimum frame numbers of each category. 
		}
		\label{fig:eachcategory}
	\end{figure}
	
	After determining object categories, we search for raw sequences of each class from YouTube\footnote{Each video is collected under the Creative Commons license.}, as it is the largest public video platform and motivates many tracking benchmarks (\eg, LaSOT~\cite{fan2020lasot}, TrackingNet~\cite{muller2018trackingnet}, GOT-10k~\cite{huang2018got} and OxUvA~\cite{valmadre2018long}). Initially, we have collected at least 30 raw videos for each class and gathered more than 600 sequences in total. Then, we carefully inspect each sequence for its availability for tracking and choose 15 sequences for each category. We verify the content of each raw sequence and remove the irrelevant parts to acquire a video clip that is suitable for tracking. We limit the number of frames in each video up to 500, which is enough for testing a tracker's performance on transparent objects, while being manageable for annotation. Eventually, TOTB consist of 225 sequences from 15 transparent object classes with 86K frames. Table~\ref{tab:totb} summarizes TOTB, and Figure~\ref{fig:eachcategory} demonstrates the average video length of each object category in TOTB.
	
	\begin{figure}[!t]
		\centering
		\includegraphics[width=0.32\linewidth]{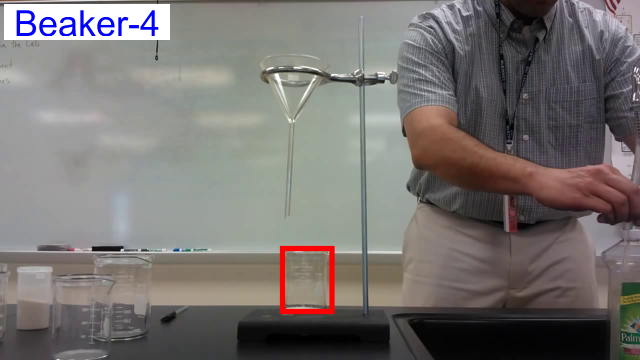}
		\includegraphics[width=0.32\linewidth]{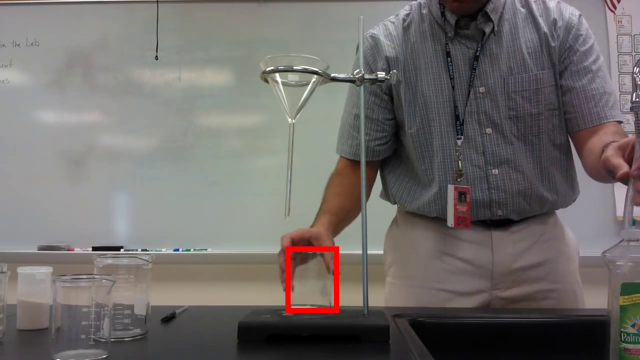}
		\includegraphics[width=0.32\linewidth]{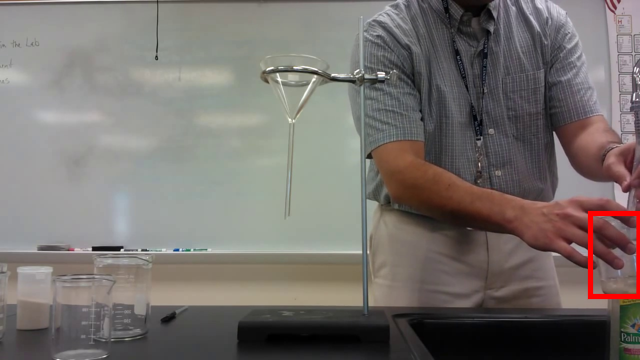}  \\
		\includegraphics[width=0.32\linewidth, height=2.1cm]{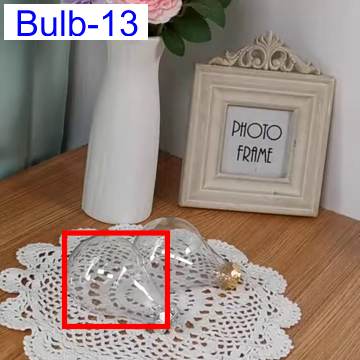}
		\includegraphics[width=0.32\linewidth, height=2.1cm]{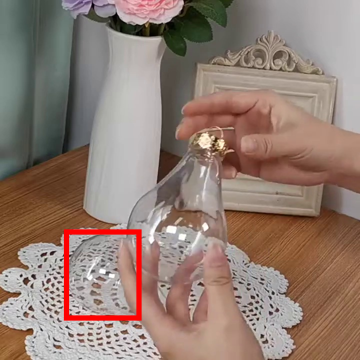}
		\includegraphics[width=0.32\linewidth, height=2.1cm]{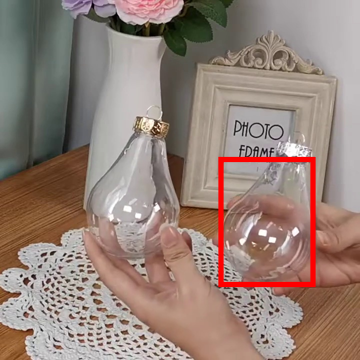}  \\
		\includegraphics[width=0.32\linewidth]{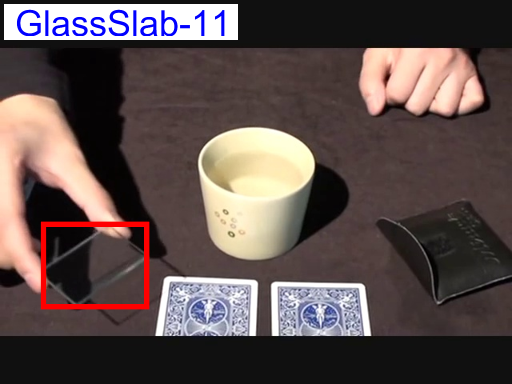}
		\includegraphics[width=0.32\linewidth]{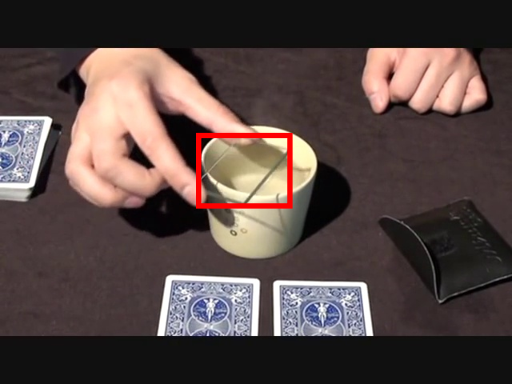}
		\includegraphics[width=0.32\linewidth]{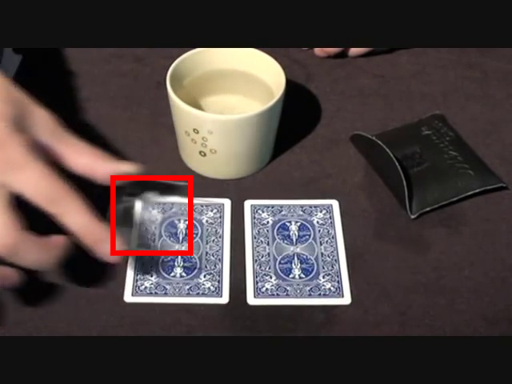}  \\
		\includegraphics[width=0.32\linewidth]{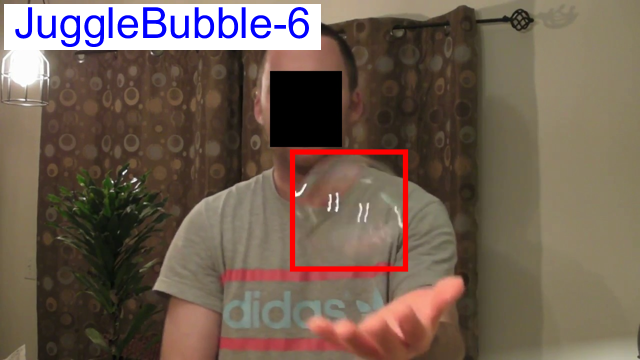}
		\includegraphics[width=0.32\linewidth]{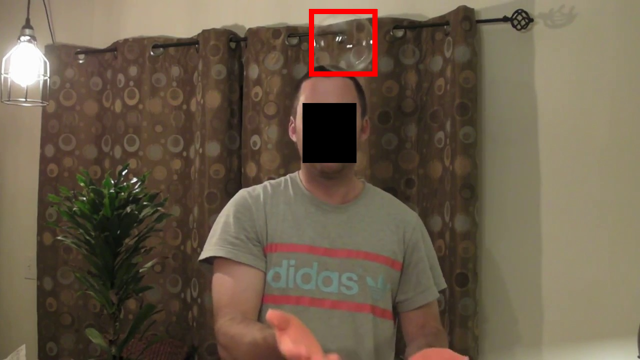}
		\includegraphics[width=0.32\linewidth]{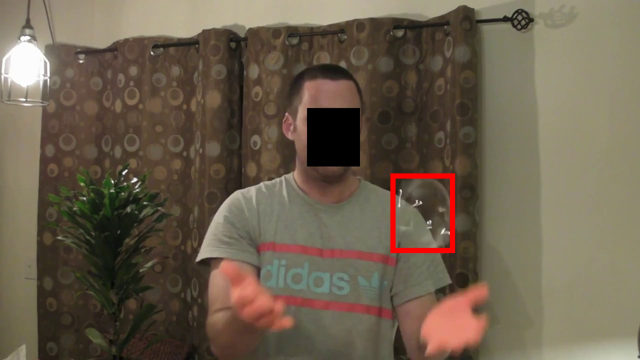} \\
		\includegraphics[width=0.32\linewidth]{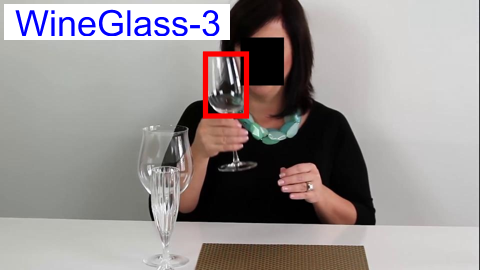}
		\includegraphics[width=0.32\linewidth]{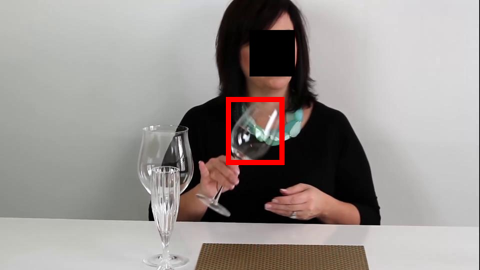}
		\includegraphics[width=0.32\linewidth]{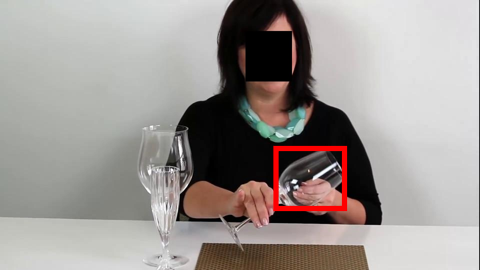} 
		\caption{Example sequences of transparent object tracking in our novel TOTB. Each sequence is annotated with axis-aligned bounding boxes.}
		\label{fig1}
	\end{figure}
	
	\subsection{Annotation}
	
	We follow the same principle as in~\cite{fan2019lasot} for sequence annotation: given the initial target in a video, for each frame, if the target appears, the annotator draws/edits an axis-aligned bounding box as the tightest one to fit any visible part of the target object; otherwise, an absence label, either \emph{full occlusion} (FOC) or \emph{out-of-view} (OV), is assigned to this frame.
	
	With the above principle, we adopt a three-step strategy for annotation, including \emph{manual labeling}, \emph{visual inspection} and \emph{box refinement}. In the first stage, each video is labeled by an expert, \ie, a graduate student who works on tacking. Since there may exist unavoidable annotation errors or inconsistencies in the first stage, a visual inspection is performed in the second stage to verify the annotation. The inspection of annotation for each video is conducted by a validation team. If the annotation result is not unanimously agreed by the members of validation team, it will be sent back to the original annotator for refinement in the third step. Such three-step strategy ensures high-quality annotation boxes for transparent objects in TOTB. Some examples for box annotation of TOTB can be found in Figure~\ref{fig1}. We show more statistics in {\textit{\supp{supplementary material}}}.

	\subsection{Attributes}
	
	Further in-depth analysis of tracking algorithms is important for researchers to understand trackers' strengths and limitations. Thus motivated, we select twelve attributes that widely exist in video tasks and annotate each sequence with these attributes, including (1) illumination variation (IV), (2) partial occlusion (POC), (3) deformation (DEF), (4) motion blur (MB), (5) rotation (ROT), (6) background clutter (BC), (7) scale variation (SV), which is assigned when the ratio of bounding box is outside the range [0.5, 2], (8) full occlusion (FOC), (9) fast motion (FM), which is assigned when the target center moves by at least 50\% of its size in last frame, (10) out-of-view (OV), (11) low resolution (LR), which is assigned when the region of the target is less than 900 pixels, and (12) aspect ratio change (ARC), which is assigned when the ratio of bounding box aspect ratio is outside the range [0.5, 2]. For each video, a 12D binary vector is provided to indicate the presence of an attribute (\ie, ``1'' denotes the presence of a certain attribute, ``0'' otherwise.)
	
	\renewcommand\arraystretch{1}
	\begin{table}[!t]\small
		\centering
		\caption{Distribution of twelve attributes on the TOTB. The diagonal (shown in \textbf{bold}) corresponds to the distribution over the entire benchmark, and each row or column presents the joint distribution for the attribute subset.}
		\begin{tabular}{R{0.55cm}C{0.18cm}C{0.18cm}C{0.18cm}C{0.18cm}C{0.18cm}C{0.18cm}C{0.18cm}C{0.18cm}C{0.18cm}C{0.18cm}C{0.18cm}C{0.18cm}}
			\toprule[1.5pt]
			& \rotatebox{90}{IV} & \rotatebox{90}{POC} & \rotatebox{90}{DEF} & \rotatebox{90}{MB} & \rotatebox{90}{ROT} & \rotatebox{90}{BC} & \rotatebox{90}{SV} & \rotatebox{90}{FOC} & \rotatebox{90}{FM} & \rotatebox{90}{OV} & \rotatebox{90}{LR} & \rotatebox{90}{ARC} \\
			\hline \hline
			IV    & \textbf{69} & 24    & 7     & 16    & 43    & 5     & 20    & 2     & 10    & 2     & 3     & 16 \\
			POC   & 24    & \textbf{110} & 18    & 38    & 59    & 23    & 48    & 9     & 26    & 7     & 12    & 40 \\
			DEF   & 7     & 18    & \textbf{42} & 6     & 6     & 8     & 24    & 0     & 7     & 0     & 1     & 20 \\
			MB    & 16    & 38    & 6     & \textbf{69} & 50    & 16    & 29    & 7     & 18    & 6     & 5     & 27 \\
			ROT   & 43    & 59    & 6     & 50    & \textbf{123} & 21    & 59    & 7     & 27    & 6     & 9     & 61 \\
			BC    & 5     & 23    & 8     & 16    & 21    & \textbf{42} & 17    & 3     & 5     & 1     & 0     & 11 \\
			SV    & 20    & 48    & 24    & 29    & 59    & 17    & \textbf{95} & 0     & 33    & 0     & 14    & 68 \\
			FOC   & 2     & 9     & 0     & 7     & 7     & 3     & 0     & \textbf{10} & 0     & 3     & 0     & 0 \\
			FM    & 10    & 26    & 7     & 18    & 27    & 5     & 33    & 0     & \textbf{44} & 0     & 11    & 29 \\
			OV    & 2     & 7     & 0     & 6     & 6     & 1     & 0     & 3     & 0     & \textbf{9} & 0     & 0 \\
			LR    & 3     & 12    & 1     & 5     & 9     & 0     & 14    & 0     & 11    & 0     & \textbf{18} & 11 \\
			ARC   & 16    & 40    & 20    & 27    & 61    & 11    & 68    & 0     & 29    & 0     & 11    & \textbf{82} \\
			\toprule[1.5pt]
		\end{tabular}%
		\label{tab:attributes}%
	\end{table}%
	
	The distribution of these attributes on TOTB is presented in Table~\ref{tab:attributes}. We can observe that the most common challenge in TOTB is \emph{rotation} (including in-place and out-plane rotations), which may cause serious feature misalignment and lead to tracking failure. In addition, the \emph{scale variation} and \emph{partial occlusion} frequently occur in videos of TOTB.
	
	
	\section{A New Baseline: TransATOM}
	\label{transatom}
	
	As mentioned early, the technical difficulty of transparent object tracking is the weak appearance caused by transparency. To address this issue, we exploit transparency feature for transparent object tracking. Specifically, considering that the transparency is a \emph{common} attribute of transparent objects, its feature should be \emph{generic} and \emph{transferable} for different transparent instances, and \emph{differentiable} from opaque objects. 
	
	Inspired by~\cite{xie2020segmenting}, we learn such transparency feature with a deep segmentation network that classifies each pixel belonging to transparent regions. Different from~\cite{xie2020segmenting} adopting a complex network, we utilize a much simpler FCN architecture~\cite{long2015fully} with ResNet-18~\cite{he2016deep} for efficient inference. The images used for training our segmentation are borrowed from the training set in~\cite{xie2020segmenting}. Note that, in our task, we only segment small and movable transparent objects. Thus, there are 2,844 static images for training. The details of the segmentation network for our task and its training are shown in \textit{\supp{supplementary material}} due to limited space.
	
	\begin{figure}
		\centering
		\includegraphics[width=\linewidth]{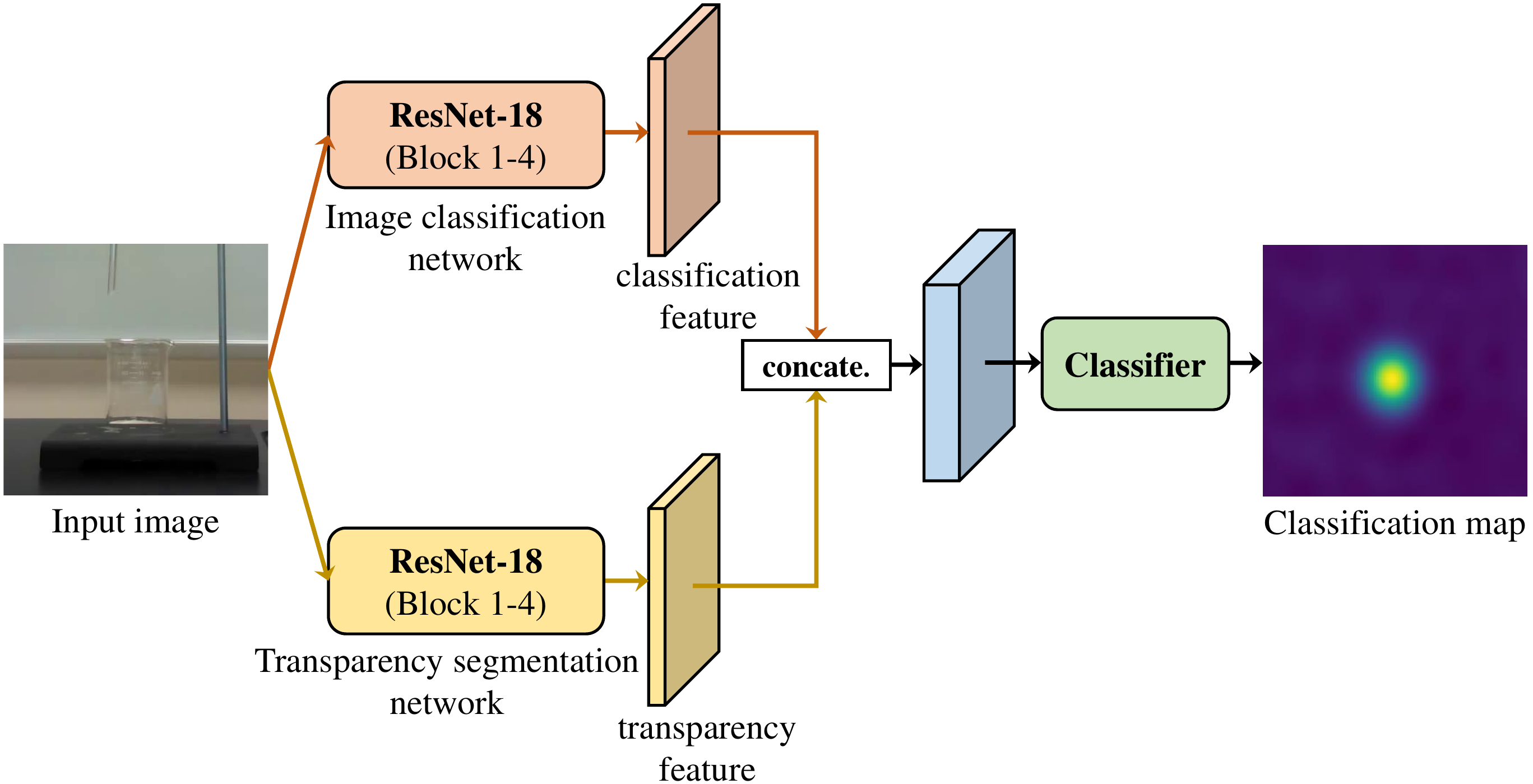}
		\caption{Illustration of architecture of TransATOM that integrates conventional classification feature and our proposed transparency feature for target localization. 
		}
		\label{fig:transatom}
	\end{figure}
	
	After training the segmentation network, we apply it for extracting transparent features for transparent objects. We integrate such feature into state-of-the-art ATOM~\cite{danelljan2019atom} to develop our new tracker, dubbed TransATOM. In particular, TransATOM consists of two feature branches. One branch is the pre-trained ResNet-18 for classification as in~\cite{danelljan2019atom}, and the other one is our trained segmentation network for transparency feature extraction. In both two branches, we extract features after block 4 and concatenate them for more robust feature representation. After that, we adopt a classification network to locate the target object. Figure~\ref{fig:transatom} shows the classification architecture of TransATOM.

	Similar to~\cite{danelljan2019atom}, the classification network consists of two convolutional layers and is formulated as follows,
	\begin{equation}
		f(X;\mathbf{w})=\phi_{2}(w_{2}*\phi_{1}(w_1*X))
	\end{equation}
	where $\mathbf{w}=\{w_1,w_2\}$ represent parameters of network and $\phi_{1}$ and $\phi_{2}$ are activation functions after each convolutional layer. $X$ is input feature to the classifier and obtained by combining both pre-trained image classification feature $x_{\mathrm{cls}}$ and transparency feature $x_{\mathrm{trs}}$ (see Figure~\ref{fig:transatom}) as follows,
	\begin{equation}
		X = x_{\mathrm{cls}} \| x_{\mathrm{trs}}
	\end{equation}
	where $\|$ denotes concatenation operation.
	
	We use the $L2$ loss to learn the classifier via
	\begin{equation}
		\ell_{\mathbf{w}} = \sum_{j=1}^{M}\gamma_{j}\|f(X_j;\mathbf{w})-Y_j\|^{2} + \sum_{k}\lambda_{k}\|w_k\|^{2}
	\end{equation}
	where $X_{j}$ is the $j$-th training sample and $Y_{j}$ is its Gaussian label centered at target location; $\gamma_{j}$ and $\lambda_{k}$ control the sample weight and the regularization amount, respectively. We use the same optimization method as in~\cite{danelljan2019atom} for learning and updating the classifier. For target scale estimation, we adopt IoU-Net as in~\cite{danelljan2019atom}. \HF{Note that, in addition to the transparency feature branch, the rest of TransATOM, including classification feature branch and IoU-Net, is directly borrowed from the baseline ATOM~\cite{danelljan2019atom}. Please refer to~\cite{danelljan2019atom} for more details.}
	
	\begin{figure}[!t]
		\centering
		\includegraphics[width=\linewidth]{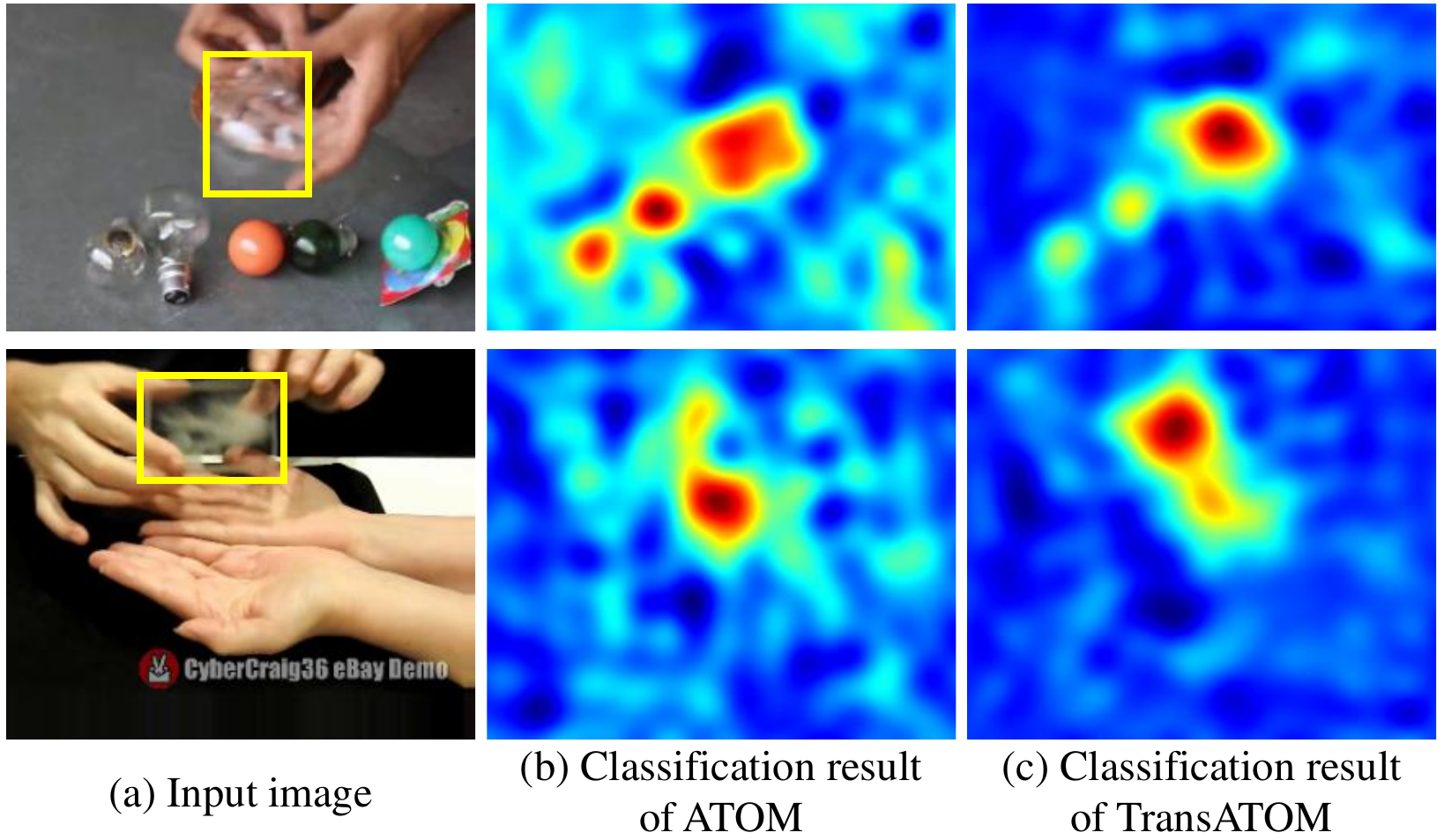}
		\caption{Classification results of ATOM and TransATOM.	We can observe that TransATOM shows better classification results for locating transparent target objects. The yellow boxes in input images are groundtruth. 
		}
		\label{fig:heatmap}
	\end{figure}
	
	Notice that, different from ATOM~\cite{danelljan2019atom}, TransATOM aims to explore additional transparency feature to improve localization of objects. Figure~\ref{fig:heatmap} shows target localization results of the two methods. We observe that TransATOM better locates the objects with the help of transparency features. Furthermore, our TransATOM runs in real-time at 26 \emph{fps}.
	
	It is worth mentioning that, the proposed transparency feature in TransATOM is {\it generic} and {\it transferable} to other trackers (\eg, DiMP~\cite{bhat2019learning} and KYS~\cite{bhat2020know}) for improvements as shown in our ablation study in Section~\ref{ablation}.

	\section{Evaluation}
	\label{res}
	
	\subsection{Evaluation Methodology}
	
	Following~\cite{fan2019lasot,muller2018trackingnet}, we use one-pass evaluation (OPE) and measure each tracker using \emph{precision}, \emph{normalized precision} and \emph{success}. The precision (PRE) measures the distance between centers of tracking results and groundtruth boxes in pixels. Different algorithms are ranked by their PRE score at a threshold (\eg, 20 pixels). To eliminate the influence of different scales, normalized precision (NPRE) is adopted by performing normalization with target areas. Success (SUC) compares the intersection over union (IoU) of tracking results and groundtruth boxes, and SUC score is computed by the percentage of tracking results whose IoU is larger than 0.5.

	\subsection{Evaluated Trackers}
	
	We evaluate 25 state-of-the-art trackers on TOTB and provide basis for future comparison. These algorithms can be roughly categorized into three types: correlation filter trackers, Siamese trackers and other deep trackers.
	
	Correlation filter tracking approaches include KCF~\cite{henriques2015high}, SRDCF~\cite{danelljan2015learning}, HCFT~\cite{ma2015hierarchical}, Staple~\cite{bertinetto2016staple}, ECOhc~\cite{danelljan2017eco}, ECO~\cite{danelljan2017eco}, STRCF~\cite{li2018learning}, StapleCA\cite{mueller2017context}, CFNet~\cite{valmadre2017end}, BACF~\cite{galoogahi2017learning} and ASRCF~\cite{dai2019visual}. The Siamese trackers consist of SiamFC~\cite{bertinetto2016fully}, SiamRPN~\cite{li2018high}, DaSiamRPN~\cite{zhu2018distractor}, C-RPN~\cite{fan2019siamese}, SPM~\cite{wang2019spm}, SiamRPN++~\cite{li2019siamrpn++}, SiamDW~\cite{zhang2019deeper} and SiamMask~\cite{wang2019fast}. For other trackers, we use MDNet~\cite{nam2016learning}, ATOM~\cite{danelljan2019atom}, DiMP~\cite{bhat2019learning}, PrDiMP~\cite{danelljan2020probabilistic}, DCFST~\cite{zhenglearning} and KYS~\cite{bhat2020know}.
	
	\subsection{Evaluation Results}
	
	\begin{figure*}
		\centering
		\includegraphics[width=\linewidth]{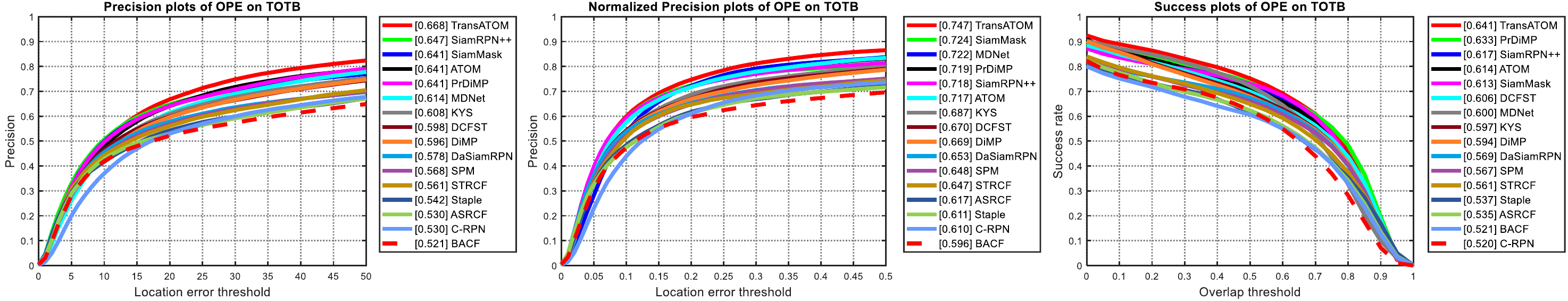}
		\caption{Tracking performance of 15 state-of-the-art trackers and TransATOM in terms of precision, normalized precision and success (please check the full results of all trackers in \emph{supplementary material}). Our TransATOM achieves the best results with all three metrics. 
		}
		\label{fig:overall}
	\end{figure*}
	
	\noindent
	{\bf Overall performance.}
	We extensively evaluate 25 tracking algorithms and our proposed TransATOM on 225 sequences in TOTB. Notice that, existing trackers are used without any modifications for evaluation. In order to avoid randomness, we run each tracker three times and average the results for its final performance. The evaluation results are reported in OPE using precision (PRE), normalized precision (NPRE) and success (SUC). Figure~\ref{fig:overall} displays the performance of 15 trackers and our TransATOM and we refer readers to the {\textit{\supp{supplementary material}}} for full results of all trackers. As demonstrated in Figure~\ref{fig:overall}. TransATOM achieves the best results with 0.668 PRE, 0.747 NPRE and 0.641 SUC. SiamRPN++ obtains the second best PRE score of 0.647, SiamMask the second best NPRE score of 0.724 and PrDiMP the second best SUC score of 0.633. In comparison with these trackers, TransATOM achieves improvements of 2.1\%, 2.3\% and 0.8\% in terms of PRE, NPRE and SUC, respectively. ATOM, which serves as the baseline of TransATOM, shows the results of 0.641 PRE, 0.717 NPRE, and 0.641 SUC. Compared to ATOM, TransATOM obtains significant performance gains of 4.1\%, 3.0\% and 2.7\%, respectively, which evidences the effectiveness and advantage of transparency feature for transparent object tracking. 

\begin{figure*}
		\centering
		\includegraphics[width=\linewidth]{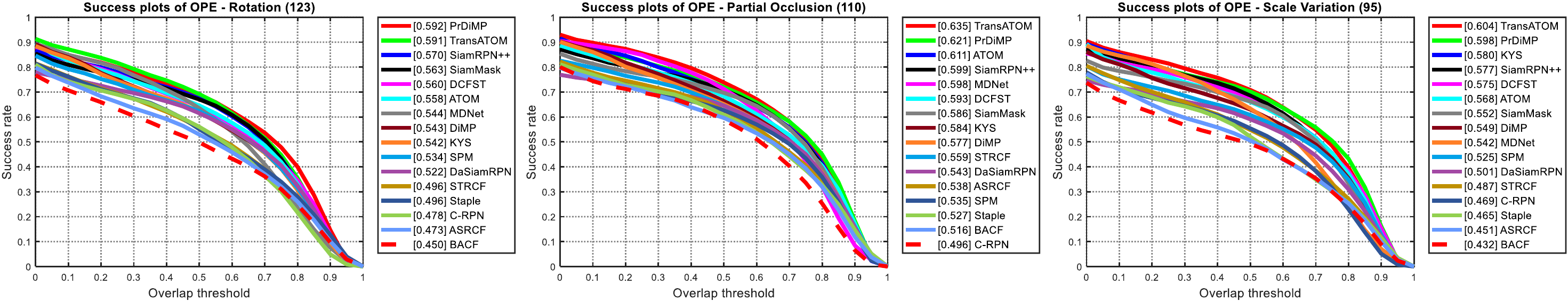}
		\caption{Tracking performance of different tracking algorithms on the three most common attributes in TOTB including \emph{rotation}, \emph{partial occlusion} and \emph{scale variation} using success (please check the full results and comparisons of all trackers in \emph{supplementary material}). 
		}
		\label{fig:att}
	\end{figure*}

	\vspace{0.5em}
	\noindent
	{\bf Attribute-based performance.} 
	In order to further analyze and understand the performance of different tracking algorithms, we conduct performance evaluation under twelve attributes. We demonstrate the results for the three most frequent challenges, including \emph{rotation}, \emph{partial occlusion} and \emph{scale variation}, in Figure~\ref{fig:att}, and refer readers to \textit{\supp{supplementary material}} for full results.
	
	\begin{figure*}[!t]
		\centering
		\begin{tabular}{@{\hspace{.0mm}}c@{\hspace{1.75mm}} @{\hspace{.0mm}}c@{\hspace{.0mm}} @{\hspace{.0mm}}c@{\hspace{.0mm}}}
			\includegraphics[width=2.7cm,height=1.7cm]{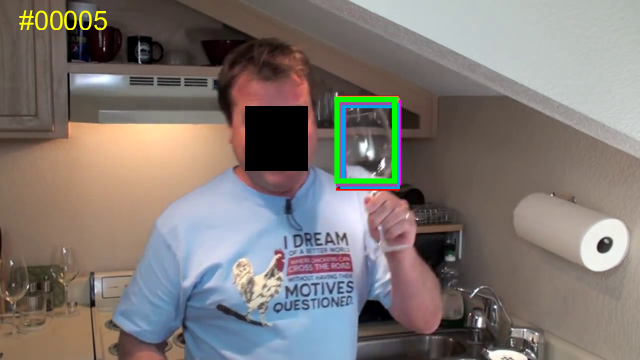} \includegraphics[width=2.7cm,height=1.7cm]{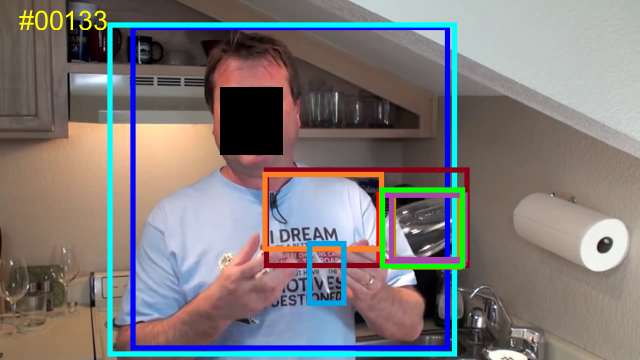} \includegraphics[width=2.7cm,height=1.7cm]{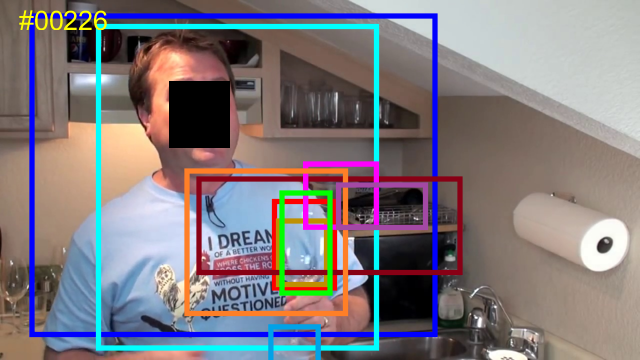} &
			\includegraphics[width=2.7cm,height=1.7cm]{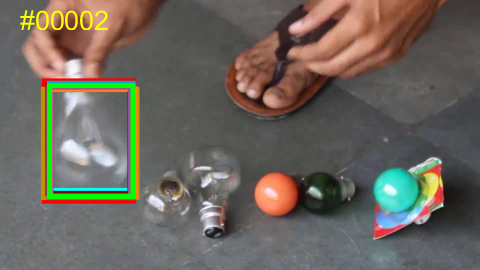} \includegraphics[width=2.7cm,height=1.7cm]{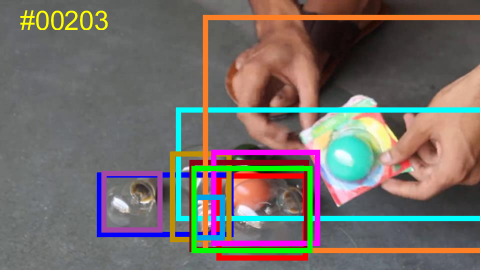} \includegraphics[width=2.7cm,height=1.7cm]{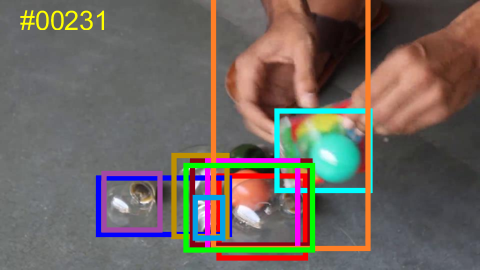}\vspace{-1mm} \\
			\vspace{1mm}\small{(a) Sequence \emph{WineGlass-7}} & \small{(b) Sequence \emph{Bulb-5}} \\
			\includegraphics[width=2.7cm,height=1.7cm]{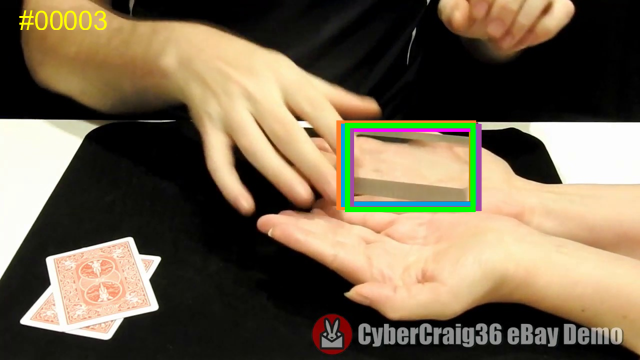} \includegraphics[width=2.7cm,height=1.7cm]{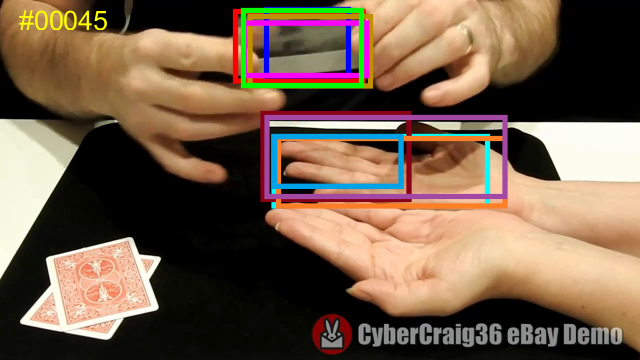} \includegraphics[width=2.7cm,height=1.7cm]{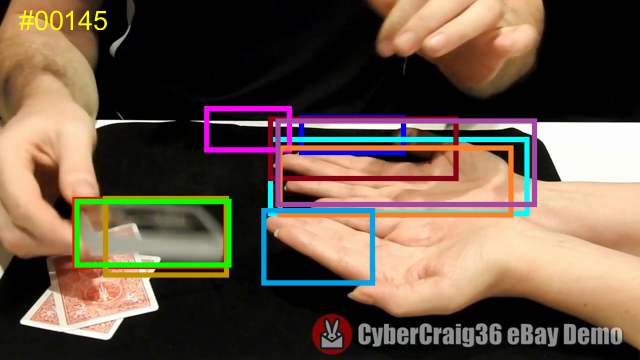} &
			\includegraphics[width=2.7cm,height=1.7cm]{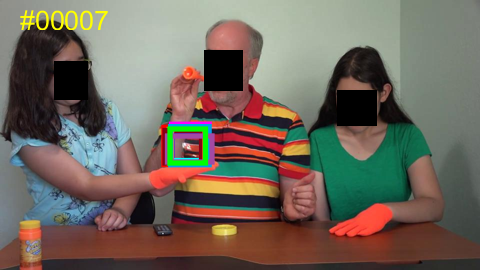} \includegraphics[width=2.7cm,height=1.7cm]{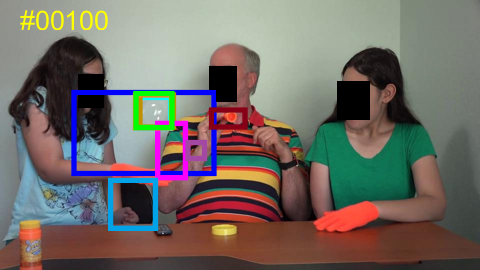} \includegraphics[width=2.7cm,height=1.7cm]{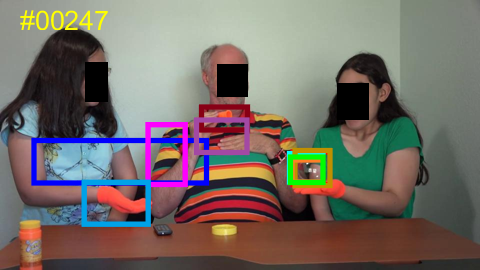}\vspace{-1mm} \\
			\vspace{1mm}\small{(c) Sequence \emph{GlassSlab-15}} & \small{(d) Sequence \emph{JuggleBubble-1}} \\
			\includegraphics[width=2.7cm,height=1.7cm]{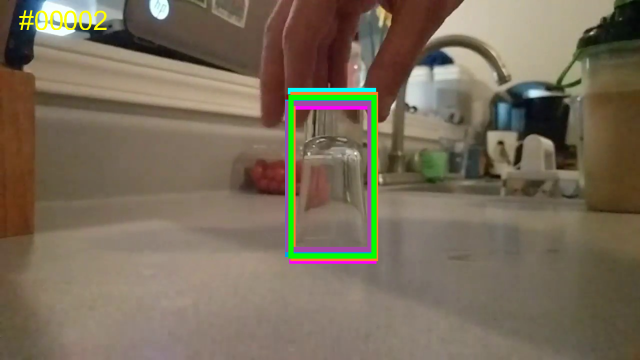} \includegraphics[width=2.7cm,height=1.7cm]{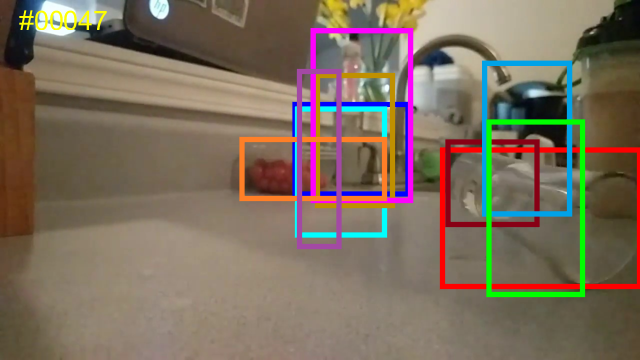} \includegraphics[width=2.7cm,height=1.7cm]{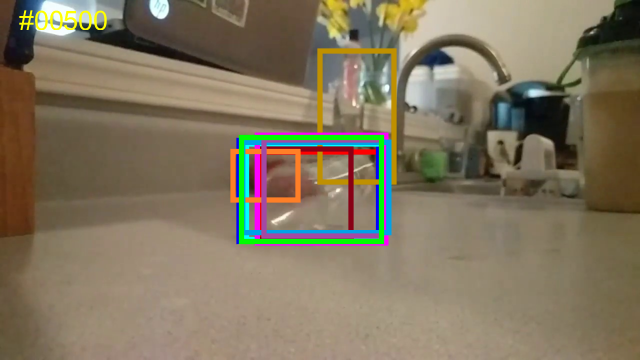} &
			\includegraphics[width=2.7cm,height=1.7cm]{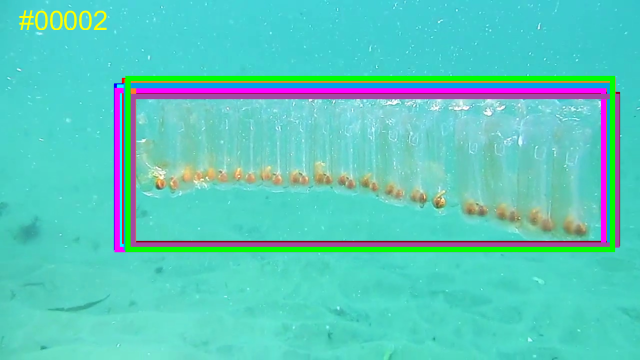} \includegraphics[width=2.7cm,height=1.7cm]{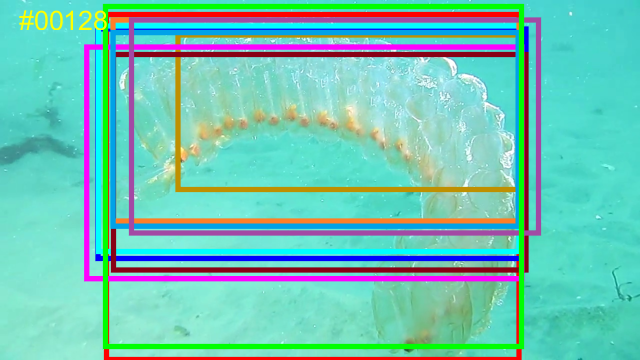} \includegraphics[width=2.7cm,height=1.7cm]{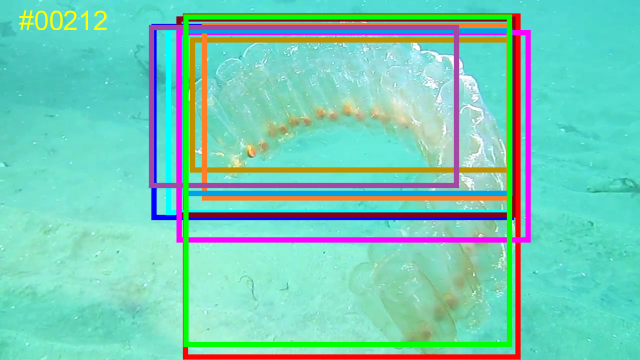}\vspace{-1mm} \\
			\vspace{1mm}\small{(e) Sequence \emph{ShotGlass-10}} & \small{(f) Sequence \emph{TransparentAnimal-11}} \\
			\multicolumn{2}{c}{\includegraphics[width=16cm]{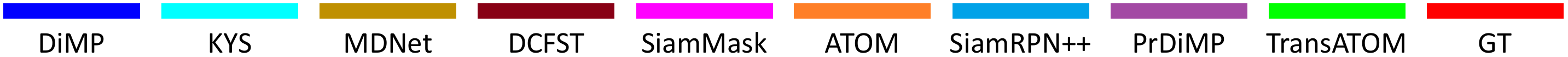}}\\
		\end{tabular}
		\caption{Qualitative results of nine trackers in six typical difficult challenges: \emph{WineGlass-7} with rotation, \emph{Bulb-5} with background clutter, \emph{GlassSlab-15} with aspect ratio change, \emph{JuggleBubble-1} with partial occlusion, \emph{ShotGlass-10} with motion blur and \emph{TransparentAnimal-11} with scale variation. The proposed TransATOM robustly locates target objects under various challenges owing to transparency feature.
		}
		\label{qual_res}
	\end{figure*}

	We observe that TransATOM performs the best on partial occlusion and scale variation. Specifically, TransATOM achieves the SUC scores of 0.635 and 0.604 on partial occlusion and scale variation, outperforming the second best PrDiMP with SCU scores of 0.621 and 0.598  by 1.4\% and 0.6\%. On the challenge of rotation, PrDiMP shows the best result with 0.592 SUC score. TransATOM ranks the second with 0.591 SUC score, which is competitive compared with PrDiMP. It is worth noticing that, PrDiMP leverages deeper ResNet-50 for feature extraction, while TransATOM adopts ResNet-18. Despite this, TransATOM shows better or competitive performance in comparison with PrDiMP owing to the effective transparent features. Besides, on all three attributes, TransATOM significantly outperforms ATOM  with SUC scores of 0.558, 0.611 and 0.568, showing the importance of transparency feature.

	\vspace{0.5em}
	\noindent
	{\bf Qualitative evaluation.}
	To better understand each tracking algorithm, we demonstrate qualitative tracking results of the top trackers, including TransATOM, PrDiMP, SiamRPN++, ATOM, SiamMask, DCFST, MDNet, KYS and DiMP, in six typical challenges consisting of \emph{rotation}, \emph{background clutter}, \emph{aspect ratio change}, \emph{partial occlusion}, \emph{motion blur} and \emph{scale variation} in Figure~\ref{qual_res}. From Figure~\ref{qual_res}, we observe that other trackers are able to deal with only one or several challenges. For example, PrDiMP performs well in dealing with aspect ratio change in \emph{GlassSlab-15} but fails in other challenges. SiamRPN++ can locate the target in \emph{ShotGlass-10} with motion blur while is prone to drift in \emph{Bulb-5} with background clutter.  MDNet works robustly in \emph{WineGlass-7} with rotation but loses the target in \emph{Bulb-5} with background clutter and \emph{ShotGlass-10} with motion blur. Similar observations are found for other trackers. Different from these methods, TransATOM well handles all challenges for robust target localization owing to the transparency features. More qualitative results can be found at the \href{https://hengfan2010.github.io/projects/TOTB/}{project website}.

	\subsection{Ablation Study}
	\label{ablation}
	
	\begin{table}[!t]
		\centering
		\caption{Analysis of different backbones for tracking performance on TOTB using SUC score. The better one is shown in \textcolor{red}{red} font.}
		\begin{tabular}{lcc}
			\toprule[1.5pt]
			& \multicolumn{1}{l}{ResNet-18} & \multicolumn{1}{l}{ResNet-50} \\
			\hline \hline
			ATOM~\cite{danelljan2019atom}  & \textcolor{red}{0.614} & 0.608 \\
			DiMP~\cite{bhat2019learning}  & \textcolor{red}{0.605} & 0.594 \\
			PrDiMP~\cite{danelljan2020probabilistic}  & \textcolor{red}{0.639} & 0.633 \\
			SiamRPN++~\cite{li2019siamrpn++} & 0.585 & \textcolor{red}{0.617} \\
			\hline
			TransATOM (ours) & \textcolor{red}{0.641} & 0.632 \\
			\toprule[1.5pt]
		\end{tabular}%
		\label{tab:backbone}%
	\end{table}%
	
	
	\vspace{0.3em}
	\noindent
	{\bf Depth of backbone.} \HF{Deep neural network has significantly improved tracking performance. In opaque object tracking, many recently proposed deep trackers using ResNet-50 as backbone significantly outperform those using ResNet-18 as backbone because of deeper features. Nevertheless, when tracking transparent objects, deeper features do {\it not} always bring performance gains. In particular, we compare four representative state-of-art trackers including ATOM, DiMP, PrDiMP and SiamRPN++ on TOTB. Table~\ref{tab:backbone} lists the comparison results using SUC scores. As displayed in Table~\ref{tab:backbone}, we observe that, when using deeper ResNet-50 as backbone, the SUC scores of ATOM, DiMP and PrDiMP are decreased from 0.614, 0.605, 0.639 to 0.608, 0.594 and 0.633, respectively. This indicates that the deeper features may hurt tracking performance for ATOM and DiMP. For SiamRPN++, when using deeper ResNet-50 as backbone, the SUC score is significantly improved from 0.585 to 0.617, showing the effectiveness of deeper features in Siamese tracker for transparent object tracking. Likewise, we conduct experiments of our tracker TransATOM using two backbones. As shown in Table~\ref{tab:backbone}, the performance of TransATOM with deeper ResNet-50 backbone is decreased in comparison with TransATOM with ResNet-18 backbone. By analyzing the impact of different backbones on tracking performance, we find that deeper features are {\it not} always beneficial for tracking of transparent objects. We hope this finding can provide a reference for transparent object tracker design in future.}
	
	\vspace{0.3em}
	\noindent
	{\bf Transparency feature.} \HF{To facilitate development of tracking algorithm on TOTB, we propose TransATOM by integrating transparency feature, which is a generic characteristic for transparent object learned explicitly, into state-of-the-art ATOM. In order to analyze the effect of transparency feature, we compare three tracking algorithms including ATOM, TransATOM-V and TransATOM. TransATOM-V is implemented by removing visual classification feature branch from TransATOM. Except for features, all other settings are the same for three trackers. Table~\ref{tab:transparency} shows the comparison results. Compared to ATOM with 0.614 SUC score, TransATOM-V obtains 0.625 SUC score with 1.1\% absolute gain, demonstrating the effectiveness of transparency feature in boosting performance. Moreover, TransATOM, which combines visual and transparency features, further pushes the performance to 0.641 and still runs in real-time.}
	
	\begin{table}[!t]\small
		\centering
		\caption{Analysis of transparency feature on tracking performance in terms of accuracy and speed.}
		\begin{tabular}{lcccc}
			\toprule[1.5pt]
			& \tabincell{c}{Visual \\ feature} & \tabincell{c}{Transparency \\ feature} & SUC & Speed\\
			\hline \hline
			ATOM~\cite{danelljan2019atom}  & \cmark     &       & 0.614 & 37 {\it fps}\\
			TransATOM-V &       & \cmark     & 0.625 & 37 {\it fps} \\
			TransATOM & \cmark     & \cmark     & 0.641  & 26 {\it fps}\\
			\toprule[1.5pt]
		\end{tabular}%
		\label{tab:transparency}%
	\end{table}%
	
	\begin{table}[!t]
		\centering
		\caption{Analysis of transferability of transparency feature.}
		\begin{tabular}{ll}
			\toprule[1.5pt]
			Trackers & SUC \\
			\hline \hline
			ATOM~\cite{danelljan2019atom}  & 0.614 \\
			TransATOM & 0.641 ($\uparrow$2.7\%) \\
			\hline
			DiMP~\cite{bhat2019learning}  & 0.594 \\
			TransDiMP & 0.613 ($\uparrow$1.9\%) \\
			\hline
			KYS~\cite{bhat2020know}   & 0.597 \\
			TransKYS & 0.619 ($\uparrow$2.2\%) \\
			\toprule[1.5pt]
		\end{tabular}%
		\label{tab:transferable}%
	\end{table}%

	\vspace{0.3em}
	\noindent
	{\bf Transferability of transparency feature.} Transparency is a {\it common} attribute of transparent objects, and transparency feature should be {\it generic} and {\it transferable}. To analyze its transferability, we integrate transparency feature into different trackers as shown Table~\ref{tab:transferable}, similar to TransATOM. We observe that, TransDiMP and TransKYS respectively improves their baseline DiMP and KYS by 1.9\% and 2.2\% gains, evidencing the transferability of transparency feature.

	\section{Conclusion}
	\label{con}
	
	In this paper, we explore a new tracking task, \ie, \emph{transparent object tracking}. In particular, we propose the TOTB, which is the first benchmark for transparent object tracking, to our best knowledge. In addition, in order to understand the performance of existing trackers and to provide baseline for future comparison, we extensively evaluate 25 state-of-the-art tracking algorithms with in-depth analysis. Furthermore, we propose a novel tracker, named TransATOM, by leveraging transparency features of transparent objects. TransATOM significantly outperforms existing state-of-the-art tracking algorithms by a clear margin. We believe that, the benchmark, evaluation and the baseline tracker will inspire and facilitate more future research and application on transparent object tracking. 
	
	\vspace{0.5em}
	\noindent
	{\bf Acknowledgment.} This work is supported in part by NSF Grant IIS-2006665 and IIS-1814745.

	{\small
		\bibliographystyle{ieee_fullname}
		\bibliography{egbib}
	}
	
	\end{document}